\documentclass[journal]{IEEEtran}
\usepackage[utf8]{inputenc}
\usepackage{graphics,epsfig,epstopdf}
\usepackage{url,amsmath,amssymb}
\usepackage{array}
\usepackage{color,subfigure}
\usepackage{booktabs}
\usepackage{threeparttable}
\usepackage{multirow}
\usepackage{tabularx}
\usepackage{verbatim}
\usepackage[table]{xcolor}
\usepackage{hyperref} 
\usepackage{booktabs}
\usepackage[noend]{algorithm}  
\usepackage[noend]{algpseudocode}  
\usepackage[numbers]{natbib}
\usepackage{amsfonts,amssymb}
\usepackage{graphicx}
\usepackage{float}
\usepackage{enumitem}
\usepackage{subfigure}
\usepackage{filecontents}
\usepackage{algorithm}
\usepackage{algpseudocode}

\usepackage{xcolor}
\usepackage{xpatch}

\begin{document}

\title{Semantic Model Component Implementation  for Model-driven Semantic Communications}

\author{Haotai Liang,
        Mengran Shi,
        Chen Dong*,
        Xiaodong Xu,~\IEEEmembership{Senior Member,~IEEE,}
        Long Liu,
        Hao Chen

\thanks{Haotai Liang and Mengran Shi are with the State Key Laboratory of Networking and Switching Technology, Beijing University of Posts and Telecommunications, Beijing, China (e-mail: lianghaotai@bupt.edu.cn; shimengran@bupt.edu.cn).}
\thanks{*Chen Dong is the corresponding author and with the State Key Laboratory of Networking and Switching Technology, Beijing University of Posts and Telecommunications, Beijing, China (e-mail: dongchen@bupt.edu.cn).}

\thanks{Xiaodong Xu is with the State Key Laboratory of Networking and Switching Technology, Beijing University of Posts and Telecommunications, Beijing, China, and also with the Department of Broadband Communication, Peng Cheng Laboratory, Shenzhen, Guangdong, China (e-mail: xuxiaodong@bupt.edu.cn).
}
\thanks{Long Liu is with the Department of Mathematics and Theories, Peng Cheng Laboratory, Shenzhen, Guangdong, China (e-mail: liul05@pcl.ac.cn).
}
\thanks{Hao Chen is with the Department of Broadband Communication, Peng Cheng Laboratory, Shenzhen, Guangdong, China (e-mail: chenh03@pcl.ac.cn).
}
}


\maketitle

\begin{abstract}
The key feature of model-driven semantic communication is the propagation of the model. The semantic model component (SMC) is designed to drive the intelligent model to transmit in the physical channel, allowing the intelligence to flow through the networks. According to the characteristics of neural networks with common and individual model parameters, this paper designs the cross-source-domain and cross-task semantic component model. Considering that the basic model is deployed on the edge node, the large server node updates the edge node by transmitting only the semantic component model to the edge node so that the edge node can handle different sources and different tasks. In addition, this paper also discusses how channel noise affects the performance of the model and proposes methods of injection noise and regularization to improve the noise resistance of the model. Experiments show that SMCs use smaller model parameters to achieve cross-source, cross-task functionality while maintaining performance and improving the model’s tolerance to noise. Finally, a component transfer-based unmanned vehicle tracking prototype was implemented to verify the feasibility of model components in practical applications.
\end{abstract}


\IEEEpeerreviewmaketitle

\section{Introduction}

\IEEEPARstart{T}{he} new 6G era of "Internet of Intelligence" with connected people, connected machines, connected things, and connected intelligence is becoming the expectation of the academia and industry \cite{56118}. Intelligence-oriented interconnection relies on semantic communication driven by Artificial Intelligence (AI) technology, which is one of the key research directions in the "Internet of Intelligence" era. By transmitting key semantic elements, it improves the efficiency of traditional data exchange in 0-1 bit streams\cite{ZHANG202260}. 

In the semantic communication network, facing different sources and tasks, the edge nodes need to use different artificial intelligence models to extract and recover the corresponding semantic information amount. However, due to the limited computing ability and storage capacity of mobile terminals, it is difficult to use large AI models with strong generalization capabilities, while AI models with a small amount of computing, memory, and storage are often only applicable to specific AI tasks and environments. So model updates are needed for edge nodes that handle multiple sources and tasks. The base station has the ability to compute, reason, and make decisions in the face of different tasks and source scenarios, supporting a "learn first, do later" mode of work mode. Therefore, model distribution from the base station to the edge nodes is one way to make the edge nodes have the ability to handle different sources and tasks.

Current semantic communication technologies and systems are only for one task or a single source\cite{9954279}, such as text \cite{9398576}, speech signal \cite{9450827}, image \cite{9791398, 9954279}, video \cite{9953110, 9955991} and 3D point cloud \cite{9931960}. 
A joint source and channel coding (JSCC) \cite{8723589} technique for wireless image transmission which directly maps the image pixel values to the complex-valued channel input symbols, encourages us to design a semantic communication system based on JSCC.
Xie and Qin designed a semantic communication system based on deep learning for text \cite{9398576} and speech signal transmission \cite{9450827}, named DeepSC and DeepSC-S, which aims at maximizing the system capacity and minimizing the semantic errors by recovering the meaning of sentences. 
Dai proposes an efficient deep joint source-channel coding method, which can closely adapt to the source distribution under nonlinear transformation, which is called nonlinear transformation source-channel coding (NTSCC) \cite{9791398}. 
The proposed NTSCC essentially learns both the latent source representation and an entropy model as the prior on the latent representation, fully exploits the semantic information of image and video, and designs the wireless image and video semantic communication system \cite{9791398, 9953110}. 
Zhu proposes an AI-powered compression and semantic-aware transmission method for point cloud video data, named AITransfer and designs an end-to-end cloud video compression and reconstruction architecture \cite{9931960}. Dynamic network conditions are combined into the end-to-end architecture design, and an adaptive control scheme based on deep reinforcement learning is adopted to provide robust transmission.

Therefore, a method is proposed to cope with the semantic extraction and recovery of edge nodes handling different sources and tasks. The concept of semantic slice model (SeSM) (To avoid confusion with the name of the network slice, we rename the semantic slice model (SeSM) as the semantic model component (SMC)) is firstly proposed in \cite{9954279} to reduce the traffic of model propagation. In the envisioned semantic communication network\cite{56118, ZHANG202260, 9954279}, the edge node sends its own semantic model update requirement, such as different tasks, and different sources to the base station according to the scenario and environment changes. The base station sends the corresponding model parameter update package, SMC, to the edge node based on the update. The edge node combines the basic model and SMC through the protocol to realize the update of its semantic model, and the edge nodes gain the ability to handle the corresponding sources and tasks.

Transfer learning was introduced to address the need for edge nodes to handle different sources and the need for model parameter updates for different tasks. In the field of transfer learning \cite{5288526}, they believe that some knowledge is specific to some domains and tasks, and some knowledge may be common between different domains, which can help improve the performance of the target domain or task. The scenario to be considered is that the target task differs from the source task, which is the inductive transfer learning scenario in the inductive transfer learning setting. 

Incremental learning is introduced to address the need for edge nodes to handle incremental tasks requiring model parameter updates. In the AI field, incremental learning has a profound research foundation for the continual learning ability of the AI model \cite{9349197}. The continual learning method based on structure expansion\cite{yan2021dynamically} freezes the feature extraction structure of the model and adds additional feature extraction structure, which makes it better to retain the features of the old task and enhance the features of the new task, which provides an essential basis and idea for the structural design of the SMC in the semantic communication network.

To address the problem that the AI model parameters are affected by channel noise during transmission at the base station and edge nodes, the reliability of the model during channel transmission is improved by adding noise to the training and fine-tuning model and making full use of the redundant information of the model to resist the model noise during channel propagation.

In this paper, two types of model components  are proposed to break through the limitations of two of them on the semantic communication system and discuss :

\begin{itemize} 
\item A semantic model component that enables the original semantic model to adapt to the other source domain is proposed to solve the limitations of special sources.
\item As time passes, AI models' unction will change with the environment. The updating SMC aims to expand the function of the base model, such as extending the classification category and adding a target detection function in the classification model. On the contrary, unloading the SMC can remove the redundant ability in the AI model.
\item We discuss how channel noise affects the performance of the model and proposes methods of injection noise and regularization to improve the noise resistance of the model.
\end{itemize}

This paper is arranged as follows: 
Section 2 introduces
the proposed SMC based semantic communication system. Section 3 details the design of the semantic model component. Experiments are presented in section 4. An implementation of SMC was presented in section 5, and conclusions about our work are drawn in section 5.
\section{SMC based Semantic Communication System}
In this section, the SMC based semantic Communication system and the details of the considered tasks are discussed. 
\subsection{System Model}
\begin{figure}
    \centering
    \setlength{\abovecaptionskip}{0.cm}
    \includegraphics[width=0.5\textwidth]{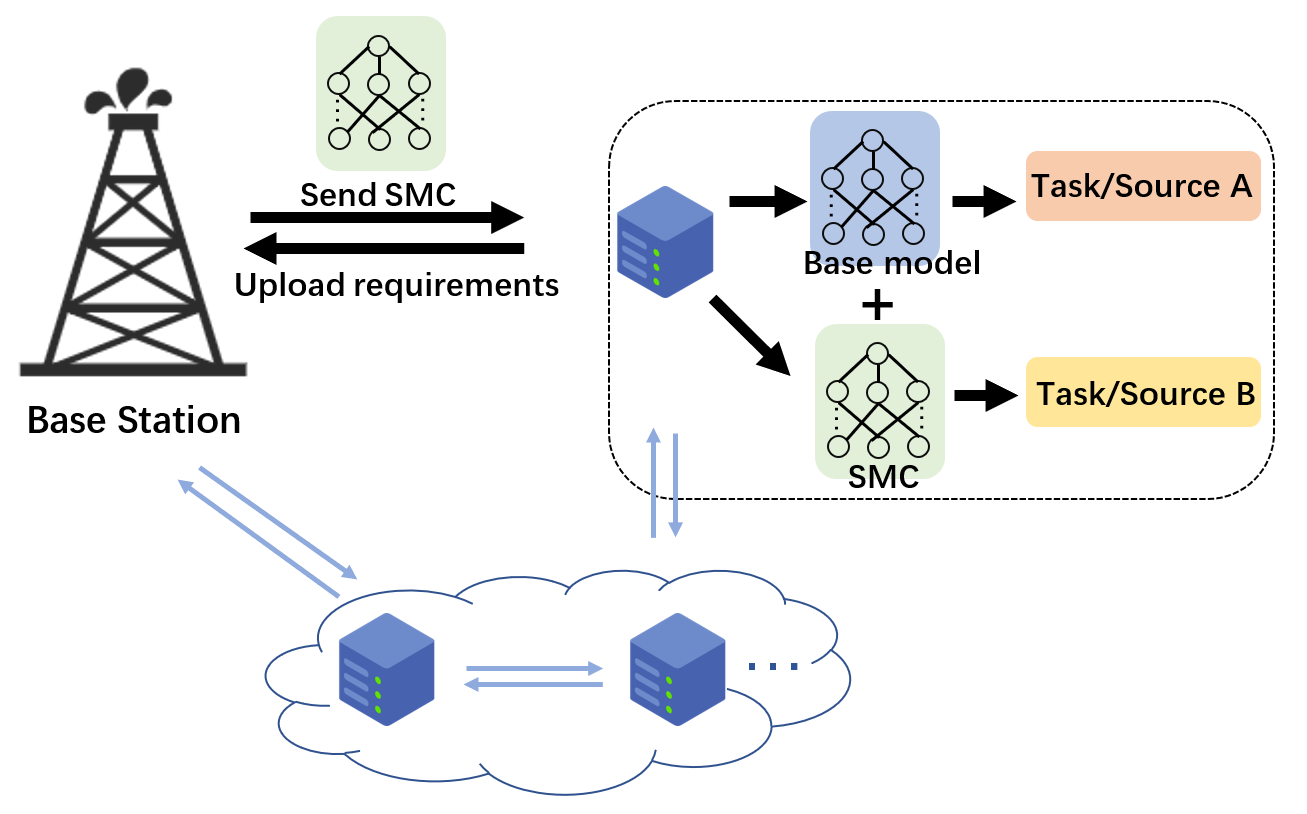}
    \centering
    \caption{Layer-based image semantic communication system consists of basic model and enhancement models. Each enhancement model can be regarded as a semantic model component (SMC) that can control accuracy and semantics. }
    \label{system_model}
\end{figure}
As shown in Fig. \ref{system_model}, edge nodes with limited memory and computing power have deployed semantic models to process specific sources and perform specific tasks. Assuming that the base station has enough computing power to train the model and enough memory to store data, while the computing power and memory of the edge nodes can only fine-tune the model and store limited data. The memory setting of the edge node is consistent with the incremental learning problem \cite{8100070}. For the incremental SMC training, the base station can observe the base model of the edge node and all data sets. The edge nodes are allowed to adopt the rehearsal strategy, which saves a part of data as the memory $\mathcal{M}$ for fine-tuning.

The intelligence transmission process based on SMC propagation can be decoupled into three sequential stages as follows.
\begin{enumerate}
\item The edge node uploads the update requirements generated by the change of the environment or the upper-level decision to the base station.
\item The base station uses enough computing power, data sets and original base models to train SMC to adapt to new source domains and tasks. Then the SMC is distributed to the edge nodes.
\item The edge node integrates the base model and SMC model so that the combined model can handle multiple source domains and tasks at the same time.
\end{enumerate}

\subsection{Task Description}
In the following, the feasibility of a SMC-based semantic communication system is demonstrated for different scenarios and tasks, and the system is analyzed from several aspects.
\subsubsection{Incremental SMC}
The purpose of incremental SMC is to expand the scope of classification or detection of the basic model. Therefore, as shown in Fig. \ref{task_description}(a), the task proposed in this paper to verify the incremental SMC is to combine the basic model with the incremental SMC to improve the categories that the model can recognize and detect. 
\subsubsection{Cross-source-domain SMC}
Cross-source domain SMC aims to expand the source distribution that the model can handle. As shown in Fig. \ref{task_description}(b), taking the semantic segmentation model as an example, the basic model and the SMC are trained using data sets with large differences in source distribution to verify the effectiveness of cross-source-domain SMC.
\subsubsection{Cross-task SMC}
\begin{figure}[h]
\centering
\begin{minipage}{0.8\linewidth}
\centerline{\includegraphics[width=\textwidth]{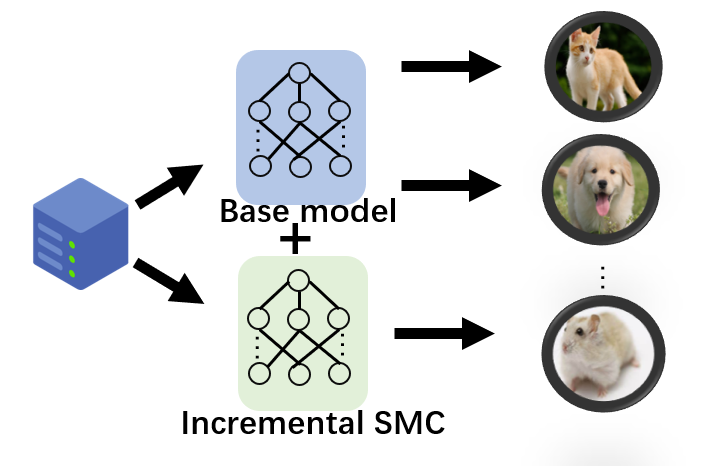}}
\centerline{(a) Incremental SMC}
\end{minipage}
\begin{minipage}{1\linewidth}
\centerline{\includegraphics[width=\textwidth]{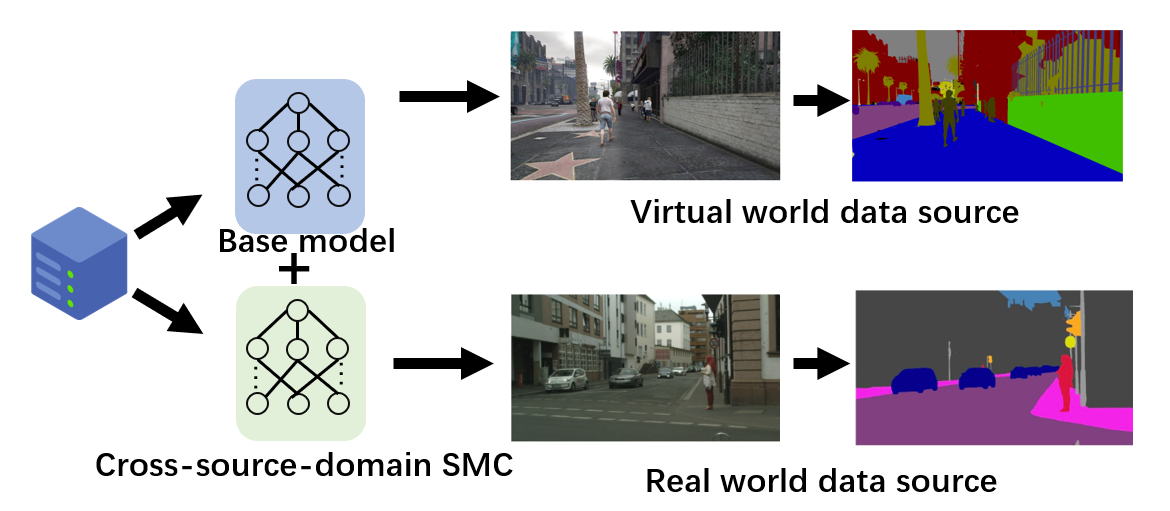}}
\centerline{(b) Cross-source-domain SMC}
\end{minipage}
\begin{minipage}{1\linewidth}
\centerline{\includegraphics[width=\textwidth]{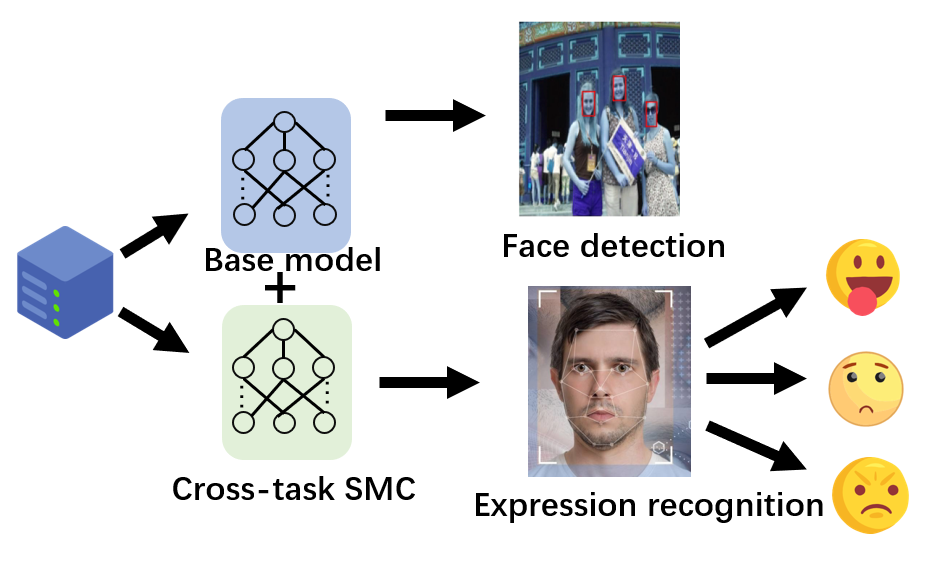}}
\centerline{(c) Cross-task SMC}
\end{minipage}
\caption{(a) Illustration of Incremental SMC framework. For image classification tasks, incremental SMC of the corresponding classification categories are added to the base model to increase the classification categories of the model. (b) Illustration of Cross-source-domain SMC framework. For the semantic segmentation task, the base model is used to perform semantic segmentation of the gta5 dataset, which is a virtual view of cars on the city streets. By adding a cross-source-domain SMC of the cityscapes dataset, the model has the ability to segment the cityscapes dataset, enabling cross-modality from virtual new data sources to real-world data sources. (c) Illustration of Cross-task SMC framework. Add cross-task SMC to the base model to give the original model the ability to perform additional tasks.}
\label{task_description}
\end{figure}

Cross-task SMC can migrate the basic model from the original function to other functions. As shown in Fig. \ref{task_description}(c), two semantic task migration SMC between classification and target detection were proposed. 

\section{Design and Implementation of the Semantic Model Component}
This section introduced the method of semantic model components, including incremental SMC, cross-source domain SMC and cross-task SMC, aiming to achieve a better trade-off between stability and plasticity. The design of the semantic model components uses the idea of structure expansion. The semantic model components trained on the base station have the module of feature extraction and task execution that the basic model on the edge node lacks. The following first describes the incremental SMC method in detail, and then extend it to cross-source domain SMC and cross-task SMC. Finally, the effect of channel noise on SMC is analyzed, and a method to improve the noise robustness is proposed.
\subsection{Method}
Firstly, the problem setup of the incremental SMC training was introduced. During the class incremental SMC training, the base station can observe the stream of the original category $\{Y\}$ and the incremental category $\{Y_{SM}\}$ and their corresponding training data $\{D\}$ and $\{D_{SM}\}$. The edge nodes are allowed to adopt the rehearsal strategy, which saves a part of data as the memory $\{M\}$ for fine-tuning.

AI models can be decomposed into feature extractor and task execution module \cite{10.5555/2969033.2969197}. The transferability between different models is mainly affected by the feature extraction module. High-level neurons will be more specific to the original task and lack generalization to other tasks, which will sacrifice the performance of the target task. 

Therefore, the first step is to find the feature extractor of the original task model that can be generalized to the target task. The part of the model with generalization will be reused, and the expanded structure will be used to improve the feature extraction ability of the target task. For the learning of the incremental SMC, the learning process can be decoupled into two sequential stages as follows.
\subsubsection{Generalized feature extractor learning}
Assuming that the base station has the models of the original task and the target task, we need to find a feature extractor that can extract common features from the two models. The Singular Vector Canonical Correlation Analysis(SVCCA) \cite{NIPS2017_dc6a7e65} method is used here to measure the similarity of the expression of the model. Given dataset $X=\{x_1, ... x_m\}$, define $z_i^l$ as the activation output of the neuron $i$ on layer $l$ of the model,
\begin{equation}
    z_i^l = \{z_i^l(x_1), ..., z_i^l(x_m)\},
\end{equation}
The representations of the $j^{th}$ layer of the original task network and the target task network is expressed by the following formula,
\begin{equation}
    l_{ori}^j = \{z_1^{l_{ori}^j}, ..., z_{m_1}^{l_{ori}^j}\},
\end{equation}
\begin{equation}
    l_{tar}^j = \{z_1^{l_{tar}^j}, ..., z_{m_2}^{l_{tar}^j}\},
\end{equation}
where $l_{ori}^{j}$ and $l_{tar}^j$ are decomposed into subspace $l_{ori}^{j'}$ and $l_{tar}^{j'}$ by singular value decomposition, which contain the most important directions to explain the variance of the original spaces. Then, the value of Canonical Correlation similarity \cite{6788402} of $l_{ori}^{j'}$ and $l_{tar}^{j'}$ is computed by projecting  $l_{ori}^{j'}$ and $l_{tar}^{j'}$ to $\tilde{l}_{ori}^{j}$ and $\tilde{l}_{tar}^{j}$,
\begin{equation}
    \tilde{l}_{ori}^{j} = W_xl_{ori}^{j'},  \tilde{l}_{tar}^j = W_yl_{tar}^{j'},
\end{equation}
The optimization objective of the Canonical Correlation Analysis(CCA) is
\begin{equation}
\underset {W_x, W_y} {\operatorname {arg\,max} } \, \frac{cov(\tilde{l}_{ori}^j, \tilde{l}_{tar}^j)}{\sqrt{D(\tilde{l}_{ori}^j)}\sqrt{D(\tilde{l}_{tar}^j)}},  
\end{equation}
where $cov(\cdot)$ and $D(\cdot)$ separately represent the covariance and variance. After obtaining the corresponding linear transformations vector ${W_x, W_y}$, we can get the correlation coefficient between $\tilde{l}_{ori}^j$ and $\tilde{l}_{tar}^j$, 
\begin{equation}
    corr(\tilde{l}_{ori}^j, \tilde{l}_{tar}^j) = \{\rho_1, ..., \rho_{min(m_1, m_2)}\},
\end{equation}
where $\rho_1$ represents the correlation coefficient of the most important direction, so $\rho_1$ was taken as the similarity between the two network layers. The higher the value of $\rho_1$, the more similar the network expression ability of this specific network layer. Generally speaking, neurons in shallow networks extract features more commonly and have correspondingly larger $\rho_1$ values.
With $\rho_T$ as the threshold, the feature extractor is divided into generalized feature extractor $\Phi_G$ and special feature extractor $\Phi_S$.

\subsubsection{Expandable representation learning}
\begin{figure*}
    \centering
    \setlength{\abovecaptionskip}{0.cm}
    \includegraphics[width=\textwidth]{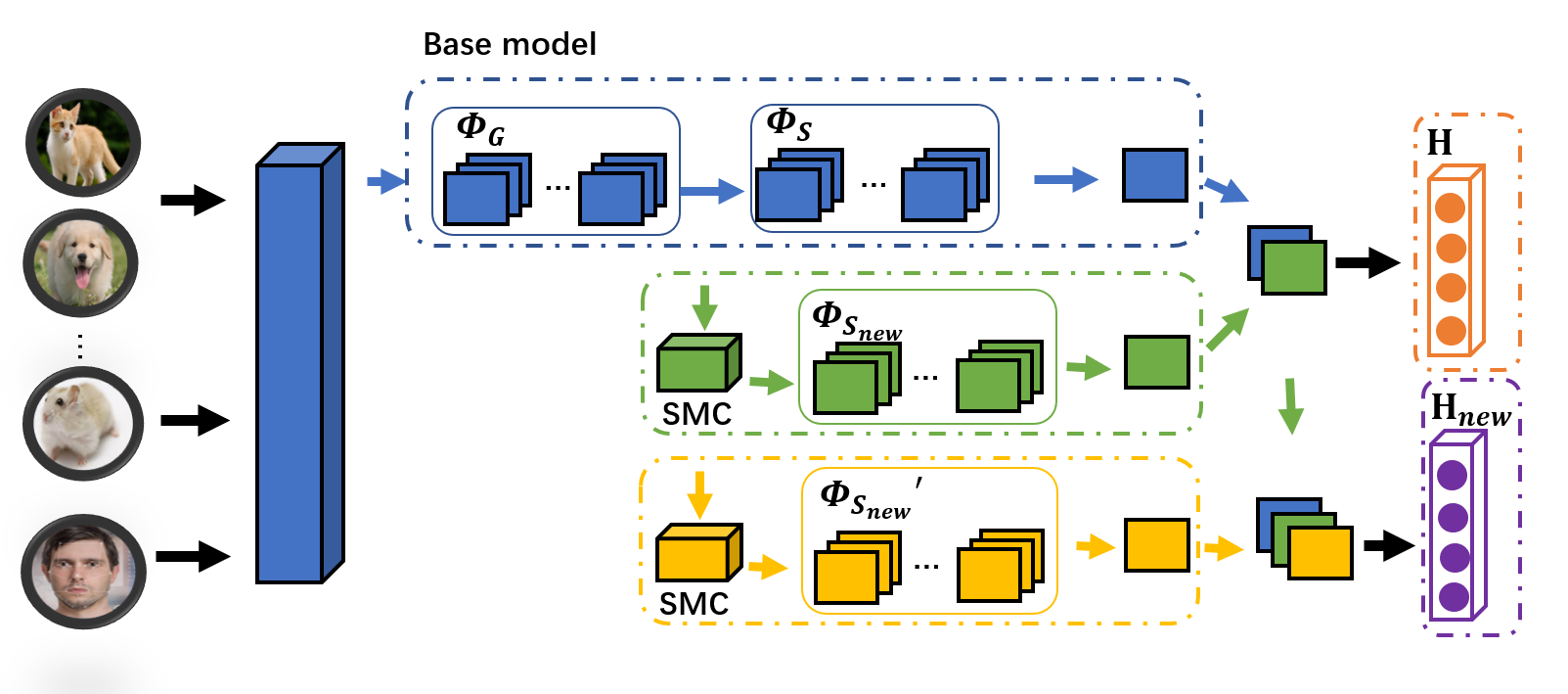}
    \centering
    \caption{Model architecture of expandable representation learning. Base model is composed of a feature extractor $\Phi = [\Phi_G, \Phi_S]$ and the task execution module $\mathcal{H}$. The generalized feature extractor $\Phi_G$ will be reused and the special feature extractor $\Phi_S$ will be expanded using the new special feature extractor  $\Phi_{S_{new}}$. In addition, the task execution model structure and parameters need to be updated for cross-task SMC.
}
    \label{paradigm}
\end{figure*}

As shown in Fig. \ref{paradigm}, our model is composed of a feature extractor $\Phi = [\Phi_G, \Phi_S]$ and task execution module $\mathcal{H}$. In the base station, the generalized feature extractor $\Phi_G$ will be reused and the special feature extractor $\Phi_S$ will be expanded. Specifically, given an image $x \in X$, updated special features are concatenated by old features and new features, 
\begin{equation}
    \Phi_S'(x) = [\Phi_S(\Phi_G(x)),  \Phi_{S_{new}}(\Phi_G(x))],
\end{equation}
where $\Phi_{S_{new}}$ represents the expanded feature extractor and is encouraged to learn the special features of new tasks.
The updated special features are fed into the task execution module to obtain the results of the new task,
\begin{equation}
    \hat{y} = \mathcal{H}{_{new}}(\Phi_S'(x)).
\end{equation}
For the learning of incremental SMC, which belongs to the incremental learning of the same task, learning the new task execution module will affect the performance of the old task. To reduce catastrophic forgetting\cite{8100070}, we freeze the parameters of the generalized feature extractor $\Phi_G$ and the special feature extractor $\Phi_S$ from the old task model, initialize the new special feature extractor $\Phi_{S_{new}}$ with the parameters of the old special feature extractor $\Phi_S$, and initialize the new task execution module $\mathcal{H}_{new}$ with the parameters of the old task execution module $\mathcal{H}$. 

The base station learns the incremental SMC with the cross-entropy loss on original and incoming data $\{D\}$ and $\{D_{SM}\}$ as follows,
\begin{equation}
    \mathcal{L}_{\mathcal{H}_{new}}=-\frac{1}{N}\sum_{i=0}^N\sum_{c=0}^My_{ic}\log{\hat{y}_{ic}},
\end{equation}
where $N$ and $M$ represent the number of samples and the number of categories respectively. To encourage the expanded special feature extractor $\Phi_{S_{new}}$ to learn the features of new tasks instead of learning the features of old tasks, the feature semantic distance \cite{10.1093/bioinformatics/btl242, 9771768} is introduced to the loss function,
\begin{equation}
    \mathcal{L}_{\Phi_S} =  \parallel\frac{1}{X}\sum_{x\in X}\Phi_{S_{new}}(x)-\frac{1}{X}\sum_{x\in X}\Phi_S(x)\parallel _2,
\end{equation}
Therefore, the incremental SMC training loss function can be expressed as follow,
\begin{equation}
    \mathcal{L}_{SM} = \mathcal{L}_{\mathcal{H}_{new}} - \lambda \mathcal{L}_{\Phi_S}.
\end{equation}
In general, SMC can be considered as part of the model with updated parameters, including the expanded feature extractor $\Phi_{S_{new}}$ and the new task execution module $\mathcal{H}_{new}$.
The algorithm for training the incremental SMC is shown in Algorithm 1.

The design idea of the Cross-source-domain SMC and Cross-task SMC  is consistent with that of the incremental SMC, but the difference is that the Cross-source-domain SMC and Cross-task SMC will not affect the old task. 

\begin{table}
\begin{center}
\begin{tabular}{p{8cm}}
\midrule
\textbf{Algorithm 1} Training Incremental SMC\\
\hline
\end{tabular}
\begin{tabular}{p{0.15cm}p{7.8cm}}
1: &\text{Initialize the classification model using base model} \\
2: &\text{Selecting classification model split points} \\
3: &\text{Add auxiliary special feature extractor $\Phi_{S_{new}}$ after the split point} \\
4: &\text{Update the task execution module $\mathcal{H} $ to $\mathcal{H}{_{new}}$} \\

5: &\text{Freeze the network parameters except $\Phi_{S_{new}}$ and $\mathcal{H}{_{new}}$} \\
6: &\text{Training $\Phi_{S_{new}}$ and $\mathcal{H}{_{new}}$ which are called the incremental SMC} \\

\hline
\end{tabular}
\end{center}
\end{table}

\subsection{Analysis of the Influence of Noise on SMC}
In the previous section, the training and application of base stations and edge servers separately were considered separately. The impact of channel noise on model transmission was ignored. Although the channel coding method can achieve lossless transmission in a better channel, it needs to add additional redundant information to the transmission information to resist noise. The model itself can be regarded as a new and special information source, which has many redundant parameters. We believe that considering the impact of channel noise from the training and fine-tuning of the model and making full use of model redundancy information will be a more elegant way to resist noise.

The robustness of the model to noise from an optimization perspective is analyzed below. AI SMC is mainly composed of convolution layers and fully connected layers, which can be expressed by the following formula,
\begin{equation}
    h(W, b) = f(Wx+b),
\end{equation}
where $f$ represents activation function, $W$ and $b$ represents the model parameters. The channel interference to the model can be considered as adding noise to the model parameters $W$ and $b$. The disturbance to the result $f(Wx+b)$ can be expressed as,
\begin{equation}
    \epsilon(W, b) = \parallel h(W, b)-h(W+N_w, b+N_b)\parallel,
\end{equation}
where $N_w$ and $N_b$ represent the noise to $W$ and $b$, respectively. Consider the first-order Taylor-expansion of $h(W, b)$ around $W$, 
\begin{equation}\label{taylor}
    h(W+N_w, b+N_b)=h(W, b)+N_wh_w'(W, b)+N_bh_b'(W, b)+R_2,
\end{equation}
where $R_2$ is the higher-order residual error of the expansion, $h_w'$ and $h_b'$ represent the gradient of $h$ to $W$ and $b$, respectively.
We hope to find the controllable parameter $C(W, b)$ larger than disturbance $\epsilon(W, b)$,
\begin{equation}\label{cc}
    \parallel h(W, b)-h(W+N_w, b+N_b)\parallel \leq C(W, b).
\end{equation}
According to formula \ref{taylor} and formula \ref{cc}, we can get
\begin{equation}\label{1}
    \parallel N_wh_w'(W, b)+N_bh_b'(W, b)\parallel \leq C(W, b),
\end{equation}
where $h_w'(W, b)$ and $h_b'(W, b)$ are computable gradients, $N_w$ and $N_b$ are obey Gaussian distribution. Therefore, formula \ref{1} can be transform as, 
\begin{equation}
    C(W, b) \approx p\sigma\parallel h_w'(W, b)+h_b'(W, b)\parallel,
\end{equation}
where $\sigma$ is the variance of the Gaussian distribution, $p$ is a constant, according to the pauta criterion, $p=3$ is taken. Therefore, minimizing the $C(W, b)$ is our optimization goal,
\begin{equation}
\begin{aligned} \label{secondorder}
&\min_{W, b} \quad p\sigma\parallel h_w'(W, b)+h_b'(W, b)\parallel\\
&\begin{array}{r@{\quad}r@{}l@{\quad}l}
s.t. &\parallel h_w'(W, b)\parallel_\infty \leq p\sigma, \\
     &\parallel h_b'(W, b)\parallel_\infty \leq p\sigma.\\
\end{array}
\end{aligned}
\end{equation}
The optimization objective encourages the model parameters to enter the noise-insensitive region, which shows the best performance. Under this model parameters, $\parallel h(W+N_w, b+N_b)\parallel$ is closest to $\parallel h(W, b)\parallel$, so the minimum point is surrounded by a flat region.

\section{Experiments and Discussions}

\begin{figure*}[h]
\centering
\begin{minipage}{0.8\linewidth}
\centerline{\includegraphics[width=\textwidth]{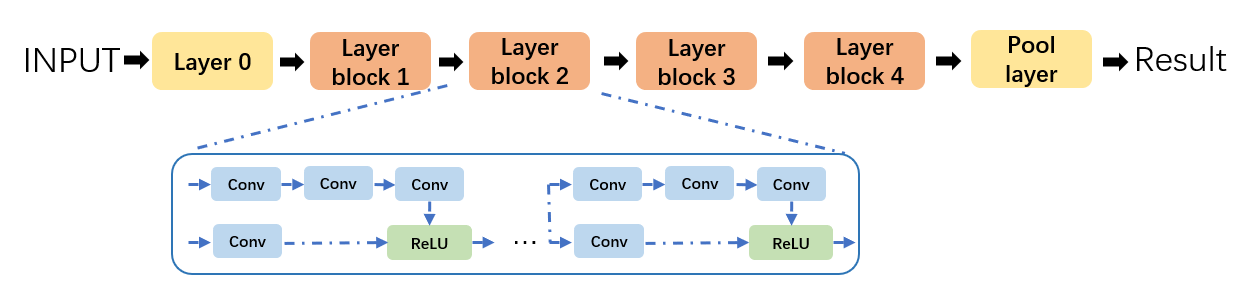}}
\centerline{(a) Incremental SMC}
\end{minipage}
\begin{minipage}{0.8\linewidth}
\centerline{\includegraphics[width=\textwidth]{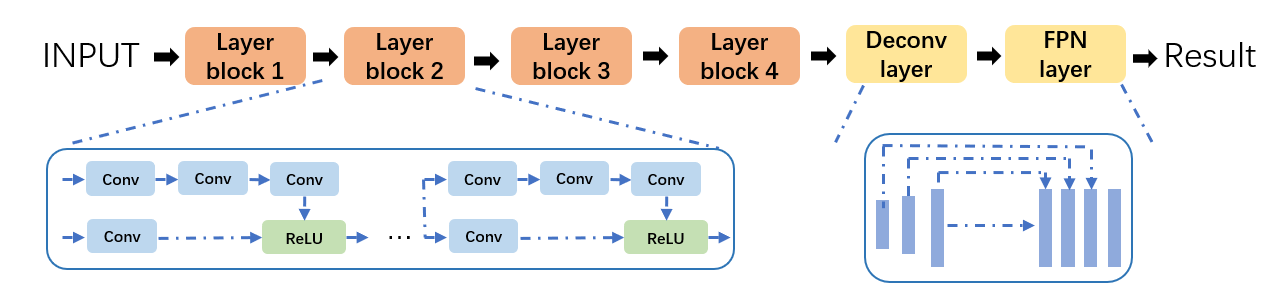}}
\centerline{(b) Cross-task SMC for Segmentation model}
\end{minipage}
\begin{minipage}{0.8\linewidth}
\centerline{\includegraphics[width=\textwidth]{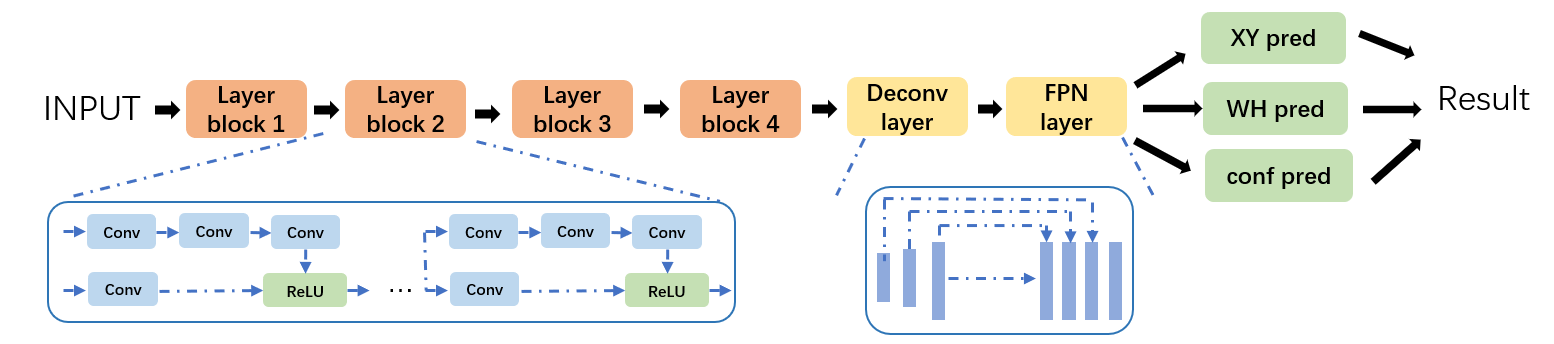}}
\centerline{(b) Cross-task SMC for Detection model}
\end{minipage}
\begin{minipage}{0.8\linewidth}
\centerline{\includegraphics[width=\textwidth]{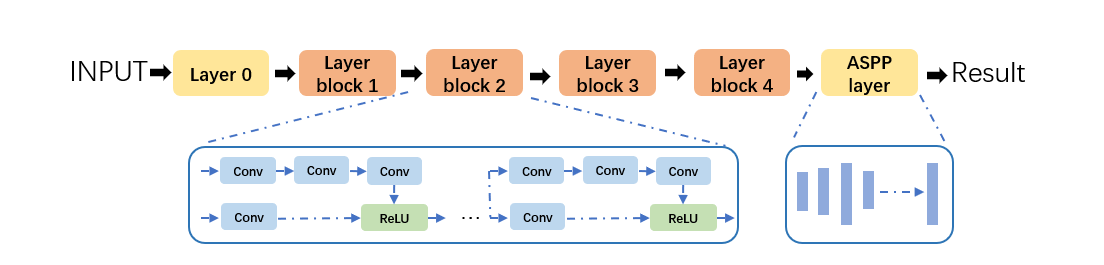}}
\centerline{(c) Cross-source-domain SMC}
\end{minipage}
\caption{(a) The structure of classification basic model and Incremental model component.  (b) The structure of Classification basic model and cross-task SMC. (c) The structure of Detection basic model and cross-task SMC. (d) The structure of Segmentation basic model and cross-source-domain SMC.}
\label{struct_description}
\end{figure*}

Three different experiments for the three types of SMC, including Incremental SMC, Cross-task SMC, and Cross-source-domain SMC mentioned above, have been designed, which are described in detail in the next three subsections. 
\subsection{Incremental SMC}
\subsubsection{Setting}
Using the incremental classification SMC as an experimental case, resnet50 is used as the basic model, and the fully connected layer is changed to a pooling layer to reduce the size of the model parameters. The basic model and the SMC are trained on the Cifar-100 dataset \cite{krizhevsky2009learning}. There are 100 classes in the Cifar-100 dataset and 600 images with the size of 32 × 32 for each class, including 500 images as training sets and 100 images as testing sets. 

First, a basic model capable of classifying 50 categories is trained, a classification model capable of classifying 60, 70, 80, 90, and 100 categories is trained as a baseline, and then incremental SMCs of 10, 20, 30, 40, and 50 are trained.

For loss function settings, $\lambda = 0.01$.
\subsubsection{Performance}
\begin{figure}
    \centering
    \setlength{\abovecaptionskip}{0.cm}
    \includegraphics[width=0.4\textwidth]{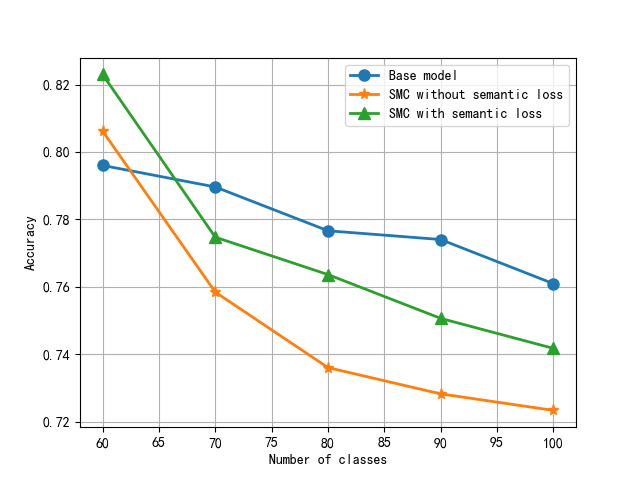}
    \centering
    \caption{The influence of the incremental SMC performance with different signal-to-noise ratio under the condition of the second-order loss regularization term.}
    \label{incre_sesm__numclass_acc}
\end{figure}

Fig. \ref{incre_sesm__numclass_acc} demonstrates the accuracy of the classification model that can classify 60, 70, 80, 90, and 100 categories and the basic model that can classify 50 categories using 10, 20, 30, 40, and 50 incremental SMC. From these results, it can be found that the classification accuracy of SMC with semantic loss is better than the classification accuracy of SMC without semantic loss. This shows that SMC can better learn incremental features using semantic loss. 
Hence, incremental SMC can achieve the processing for added tasks. 
Although the accuracy of the base model with incremental SMC is not as good as the model with 60, 70, 80, 90, and 100 classes of classification itself, the use of incremental SMC can avoid the propagation of the entire model parameters and effectively reduce the model transmission bandwidth. 
In Fig. \ref{CCA} (a), the reason that the performance of the model using SMC is better than the base model for 60-class classification is incremental structure makes the semantic model components larger in parameter size than the base model.

Fig. \ref{struct_description} (a) shows the structure of the classification model, which consists of layer blocks, also known as the bottleneck in the Resnet50. The model is partitioned into multiple bottlenecks, and the average CCA of each bottleneck is calculated to obtain CCA metrics for different split points.

From Fig. \ref{CCA} (a), it can be seen as the network structure deepens, the CCA value gradually decreases. According to the meaning of the CCA representation, it indicates that the classification capability of the network which uses SMC with 50 incremental classes and the capability of the network which uses SMC with 30 incremental classes are more similar at the shallow network level. 
For the SMC with the same incremental setting, i.e., with the same 10, 30, and 50 classification capacity, SMCs with different parameter sizes were prepared by selecting different split points according to the structure of the classification network. When the split points are shallower, the larger range SMC is applied in the network. Correspondingly, the SMC parameter size becomes larger, so the higher the classification accuracy obtained.
In addition, when the split point is located before layer block 4, as the split point position deepens, the SMC parameter size decreases, and the accuracy rate decreases slowly. When the split point is located after layer block 4, the accuracy rate decreases rapidly. 
Therefore, considering the comprehensive performance and transmission efficiency, setting the split point at layer block 4 can obtain a higher accuracy rate and save more model transmission bandwidth.

Table. \ref{vit incrematal SeSM} shows the results of using Vision transformer (ViT) as the base model for image classification and using incremental SMC in ViT. As can be seen, similar to the results of image classification using Resnet50, the accuracy of using SMC on the base model is similar to the accuracy of a model that itself has incremental category classification capability, but the propagation of the entire model parameters of 14.34 MB can be avoided by using the SMC approach. It also demonstrates that the SMC method can be applied to transformer structures.

\begin{table}[]
\caption{Accuracy and the parameter size of base model and different incremental SMCs}
\begin{tabular}{|c|cccccc|}
\hline
{\color[HTML]{000000} \textbf{Number of classes}} & \multicolumn{1}{c|}{{\color[HTML]{000000} \textbf{60}}} & \multicolumn{1}{c|}{{\color[HTML]{000000} \textbf{70}}} & \multicolumn{1}{c|}{{\color[HTML]{000000} \textbf{80}}} & \multicolumn{1}{c|}{{\color[HTML]{000000} \textbf{90}}} & \multicolumn{1}{c|}{{\color[HTML]{000000} \textbf{100}}} & {\color[HTML]{000000} \textbf{\begin{tabular}[c]{@{}c@{}}Parameter\\    \\ Size\end{tabular}}} \\ \hline
{\color[HTML]{000000} Base model}                 & \multicolumn{5}{c|}{{\color[HTML]{000000} 0.966}}                                              & {\color[HTML]{000000} 14.34MB}                                                                 \\ \hline
{\color[HTML]{000000} SMC}                       & \multicolumn{1}{c|}{{\color[HTML]{000000} 0.9229}}      & \multicolumn{4}{c|}{{\color[HTML]{000000} 0.9208} }  & \multicolumn{1}{c|}{{\color[HTML]{000000} 6.03MB} }                                  \\ \hline
\end{tabular}
\label{vit incrematal SeSM}
\end{table}

\subsection{Cross-task SMC}
\subsubsection{Setting}
In the experiment, cross-task SMC becomes the bridge between the target detection task and the semantic segmentation task. The basic model and the SMC are trained on the Pennfudan dataset \cite{wang2007object}, which has 170 images with 345 labeled pedestrians for pedestrian detection and segmentation.
\subsubsection{Performance}
Taking the two tasks of image segmentation task and target detection task as an example, we add the SMC of image segmentation to the target detection model to make the model have the capability of segmentation. Similarly, the target detection SMC is added to the image segmentation model to enable it to have the target detection ability. 

U-net \cite {ronneberger2015u} is used as the base model for image segmentation in the experiments, and the target detection model adds a target detection head to U-net. 
The basic architecture is referenced from the anchorless target detection model \cite {9010985}. The anchor-free model transforms target detection into a keypoint detection problem without clustering multiple wide-height anchor parameters on the current training data prior to training by keypoint detection or localizing the centroid of the target object.

Fig. \ref{struct_description}(b) shows the structure of the Segmentation SMC model. 
Using segmentation SMC equips the detection model with the ability to perform the segmentation task. The network is artificially divided into five parts: block 1 layer, block 2 layer, block 3 layer, deconv layer, and FPN layer. We set the split points to calculate the average CCA of each part separately to obtain the parameter similarity of different parts of the network between the target detection task and the image segmentation task. 
The metric selected here to measure the target detection effectiveness is IoU, which is calculated as follows:
\begin{equation}
    IoU=\frac{Area\ of\ Intersection\ of\ two\ boxes}{Area\ of\ Union\ of\ two\ boxes} 
\end{equation}

Fig. \ref{CCA}(b) demonstrates the IoU results with different split points. 
A model with image segmentation capability is used as the base model, and SMCs with image segmentation are added to the original model for target detection to compare performance. 
For the detection model, when the split point is located in the FPN layer, the corresponding IoU value is close to 0. It indicates that the original target detection model is little to no capability for image segmentation without using SMC. 
As the split point gradually moves forward in the network structure, the image segmentation SMCs are gradually expanded in the network, so the image segmentation performance is improved. When the split point is at block 1, the IoU value of the segmentation task is almost the same as the IoU result of the base model. 
It can be seen that the deeper the network level, the poorer the correlation between network parameters, while the shallow network parameters have a higher similarity. So when we use cross-task SMC in deeper network levels, a larger performance improvement will be achieved. This provides a trade-off between stability and plasticity. Therefore, a better choice of the split point is located at block 3, which has higher performance and saves more bandwidth for model propagation.

Fig \ref{struct_description} (c) is the structure of the detection SMC model. Based on the anchorless model, the target detection SMC was added to the U-net model to make the model have target detection ability. 
Split points are set at block 1 layer, block 2 layer, block 3 layer, deconv layer and FPN layer to obtain the average value of CCA for the above network structure and use SMCs at different split points to enable target detection. 
AP 50 and AP 75 are used as a measure of target detection performance; AP 50 means that if the IoU value between the predicted and real boxes is greater than 0.5, the prediction is considered correct, and less than 0.5 is considered incorrect. AP 75, similarly, if the IoU value between the predicted and real boxes is greater than 0.75, the prediction is correct, otherwise it is incorrect.

As shown in Fig \ref{CCA} (c), as the split point is moved forward, the SMC parameter size becomes larger. Correspondingly, the range of SMCs applied in the network expands, so the model performance gains. 
When the split point is located before block 3, the SMC parameter size decreases. Hence the model performance decreases slowly. When the split point is located after block 3, the model performance decreases rapidly. Considering the transmission efficiency of model propagation and model performance, the best choice is to choose the split point at block 3.


\subsection{Cross-source-domain SMC}
\subsubsection{Setting}
In the experiment, the source SMC overcomes the limitations of domains between different datasets. Take the semantic segmentation task as an example, using deeplab \cite{chen2017rethinking} as the basic model. The basic model firstly was trained on the gta5 dataset \cite{richter2016playing}, which contains 24966 synthetic images with pixel-level semantic annotation. Then the source SMC was trained on the Cityscapes dataset \cite{Cordts2016Cityscapes}.  

\subsubsection{Performance}
Fig. \ref{struct_description} (d) shows the structure of the Source SMC model. The deeplab semantic segmentation model is divided into two parts, encoder and decoder. The encoder part consists of resnet18, and the decoder part uses ASPP. The split points are set at layer1, layer2, layer3, and layer4 in the deeplab network, the values of CCA were calculated for each network component above to obtain the similarity between the network parameters applicable to the gta5 dataset and the Cityscapes dataset. The semantic segmentation performance of the network on the Cityscapes dataset is calculated by using SMCs of different sizes, and the mIoU is used as the evaluation metric, which is expressed as the average of the intersection ratio of each class in the dataset. 

\begin{figure}
\centering
\begin{minipage}{0.8\linewidth}
\centerline{\includegraphics[width=\textwidth]{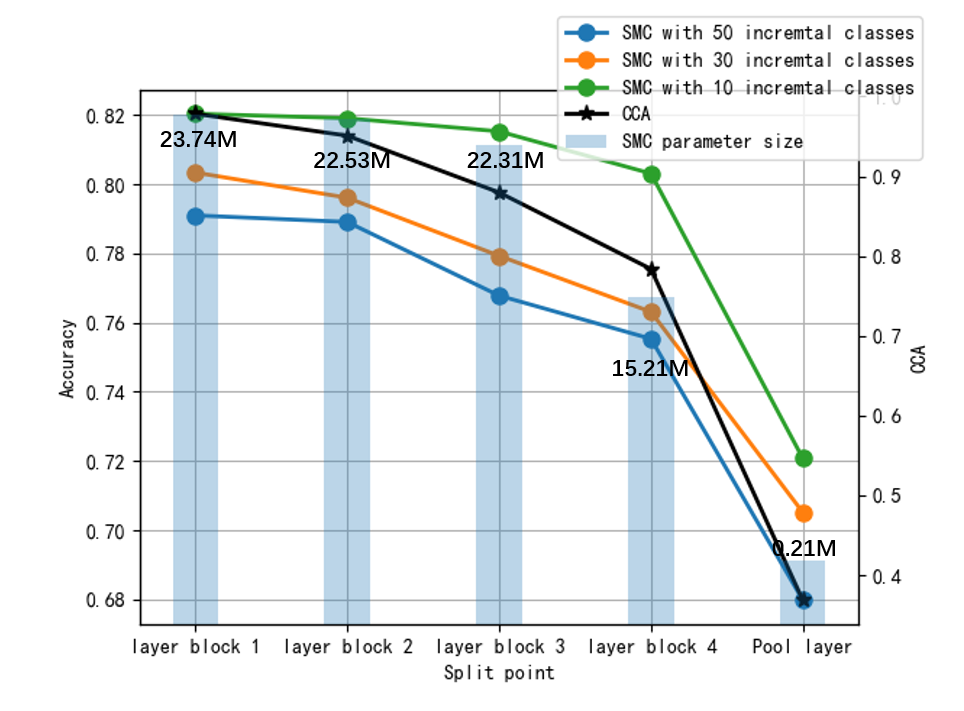}}
\centerline{(a) Incremental SMC}
\end{minipage}
\begin{minipage}{0.8\linewidth}
\centerline{\includegraphics[width=\textwidth]{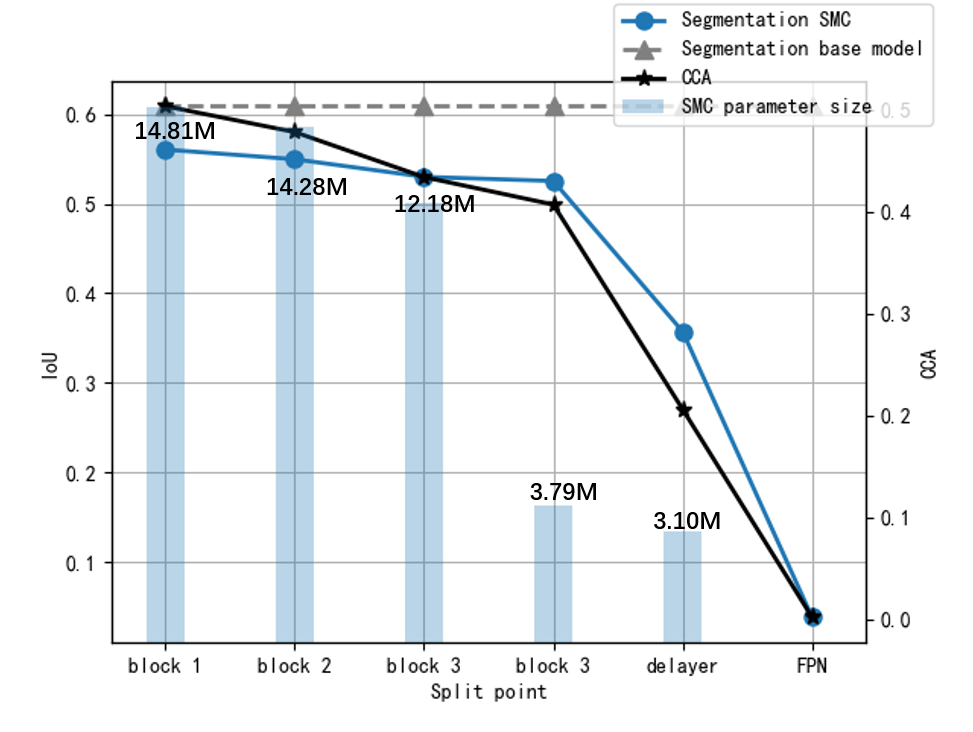}}
\centerline{(b) Cross-task SMC for Segmentation model}
\end{minipage}
\begin{minipage}{0.8\linewidth}
\centerline{\includegraphics[width=\textwidth]{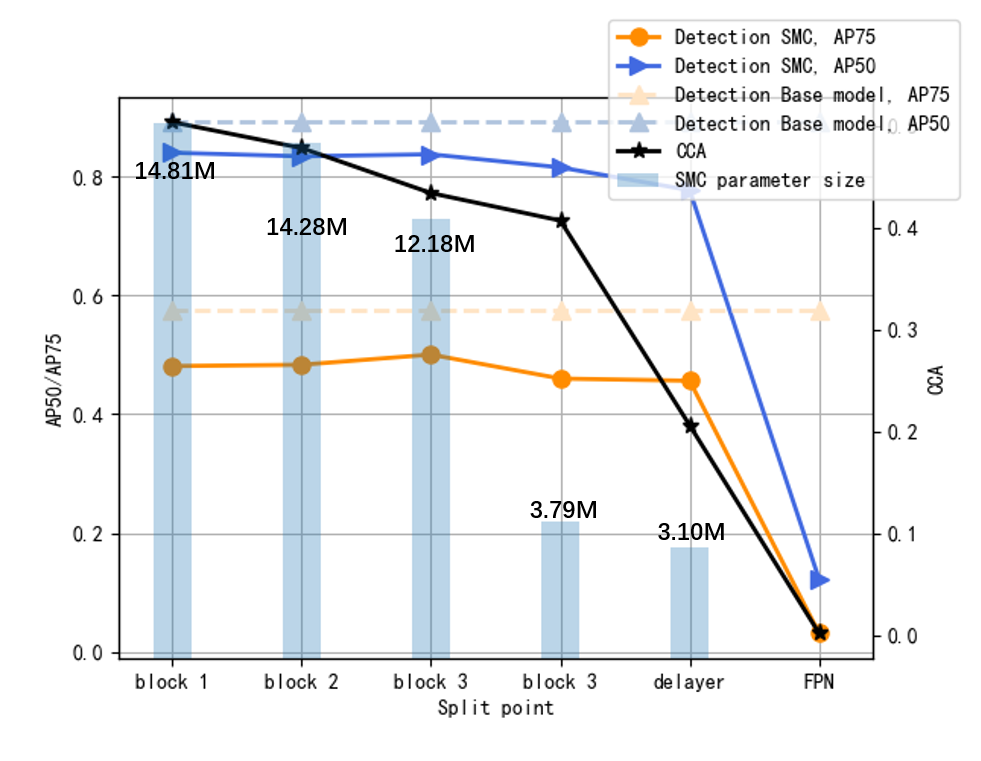}}
\centerline{(c) Cross-task SMC for Detection model}
\end{minipage}
\begin{minipage}{0.8\linewidth}
\centerline{\includegraphics[width=\textwidth]{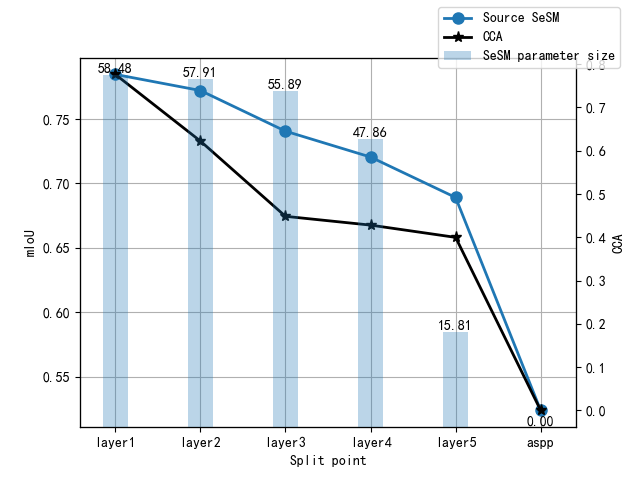}}
\centerline{(d) Cross-source-domain SMC}
\end{minipage}
\caption{(a) CCA, size and accuracy of corresponding components at different split points. (b) CCA, size and IoU of corresponding components at different split points. (c) CCA, size, AP50 and AP75 of corresponding components at different split points. (d) CCA, size and mIoU of corresponding components at different split points.}
\label{CCA}
\end{figure}

Fig. \ref{CCA} (d) demonstrates the mIoU results among different split points. The results trained on the gta5 dataset do not apply to Cityscapes dataset if the source SMC is not used, which illustrates the different domains of different datasets. 
As the split point gradually moves forward, the source SMC parameter size gradually becomes larger, and correspondingly the mIoU value of semantic segmentation in the Cityscapes dataset becomes higher. 
When the split point is moved forward from the aspp layer to layer 5, the SMC parameter size increases, and the semantic segment performance gets a substantial improvement, and when it is moved forward again, the performance improvement is gradually slow. 
This provides a trade-off between stability and plasticity. 
Therefore, a better choice of the split point is located at layer 5, which has higher performance and saves more bandwidth for model propagation. 
It is observed that the parameters in deeper layers tend to exhibit higher variance than those in shallow layers, as the former is responsible for capturing more intricate features compared to the latter. 
Therefore, by only fine-tuning the deeper layers, we can effectively adapt the network to new sources of data without having to retrain the entire network. 
This approach can be particularly beneficial in scenarios where the available bandwidth for model propagation is limited, such as in mobile or edge computing applications.

\subsection{Impact of the noise on model}
\subsubsection{Setting}
In this section, SMC with 10 incremental classes is used as an example to analyze the effect of channel noise on SMC performance. The second-order loss regularization term from the formula \ref{secondorder} is added to the training loss function and set $\beta$ as the loss trade-off term.
The total loss function can be re-described as follows,
\begin{equation}
    \mathcal{L}_{SM} = \mathcal{L}_{\mathcal{H}_{new}} - \lambda \mathcal{L}_{\Phi_S} - \beta\parallel h_w'(W, b)+h_b'(W, b)\parallel.
\end{equation}
\subsubsection{Performance}
\begin{figure}
    \centering
    \setlength{\abovecaptionskip}{0.cm}
    \includegraphics[width=0.4\textwidth]{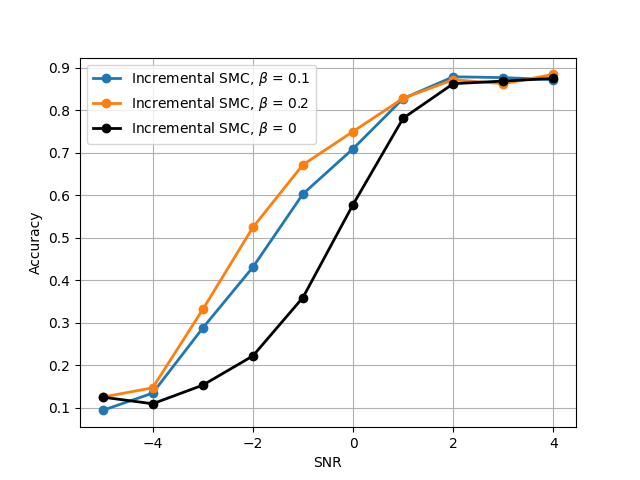}
    \centering
    \caption{The influence of the incremental SMC performance with different signal-to-noise ratio under the condition of the second-order loss regularization term.}
    \label{system}
\end{figure}

The accuracy of image classification at different signal-to-noise ratios was tested by adding noise to 10 incremental SMCs. As shown in Fig. \ref{system}, it can be known that at low signal-to-noise ratios, the larger the beta value, i.e., the greater the proportion of the loss function used to resist the noise component in the training process, so the higher the image classification accuracy. 
At SNR of -1 and -2, using a beta equal to 0.2 is about 3 dB more accurate than classifying images without anti-noise SMC. In contrast, at higher signal-to-noise ratios, the accuracy of the noise-resistant model is essentially the same as that of the model without the noise-resistant SMC when the signal-to-noise ratio is higher than 2. This shows that our model can not affect the model performance under high SNR conditions and effectively resist noise under low SNR conditions.

\section{Prototype and Implementations}
In this section, incremental SMC is applied to unmanned vehicle identification systems, aiming to enlighten and solve the problem of rapid identification of unknown targets by unmanned vehicle systems.

\subsection{Prototype System}
\begin{figure}
    \centering
    \setlength{\abovecaptionskip}{0.cm}
    \includegraphics[width=0.4\textwidth]{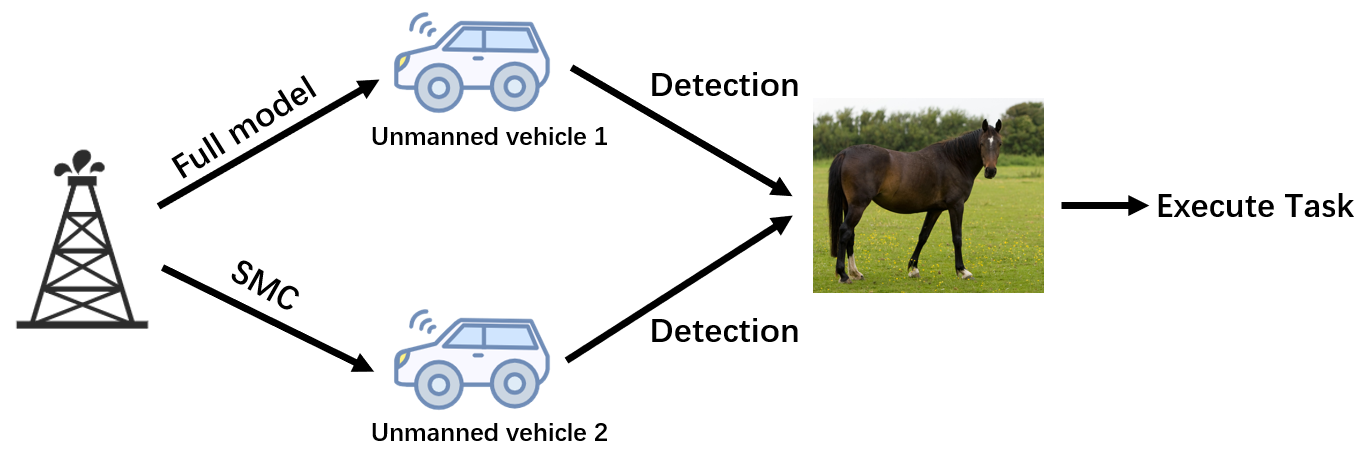}
    \centering
    \caption{Prototype comparison scheme. The edge server transmits the full model and the components to the unmanned vehicles 1 and 2, respectively. After receiving the model and the component, they detect the new targets, and then performs subsequent tasks.}
    \label{Prototype system}
\end{figure}

The experimental scheme of the unmanned vehicle identification system is shown in Fig. \ref{Prototype system}. The prototype system consists of two unmanned vehicles and an edge server. Unmanned vehicles perform target detection and tracking tasks. When unmanned vehicles encounter unrecognized targets, the edge server sends the entire model and the incremental SMC that can identify new targets to unmanned Vehicle 1 and unmanned Vehicle 2, respectively. Unmanned vehicle 1 and unmanned vehicle 2, respectively utilize the received entire models and the SMC  to adapt to the dynamic changes of the mission target, reidentify and continue tracking the target. 

\subsection{Hardware Platform}
\begin{figure}
\centering
\begin{minipage}{0.8\linewidth}
\centerline{\includegraphics[width=\textwidth]{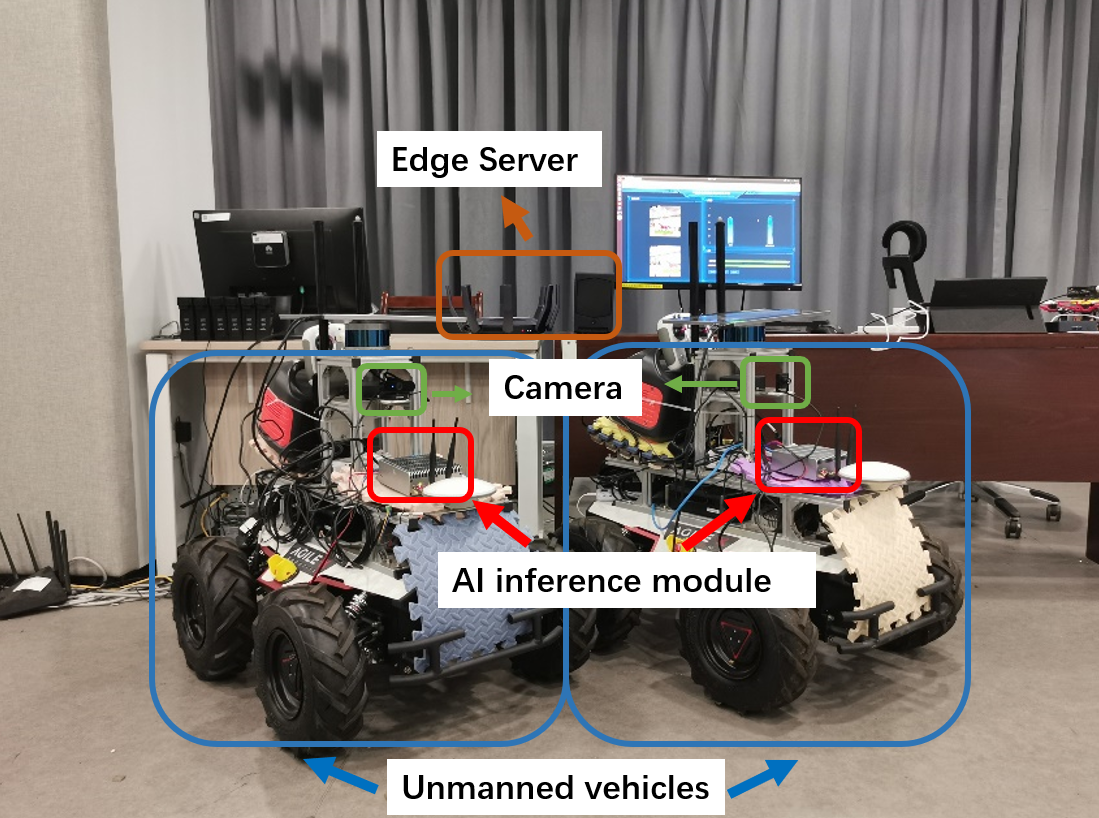}}
\centerline{(a) Hardware platform}
\end{minipage}
\begin{minipage}{0.8\linewidth}
\centerline{\includegraphics[width=\textwidth]{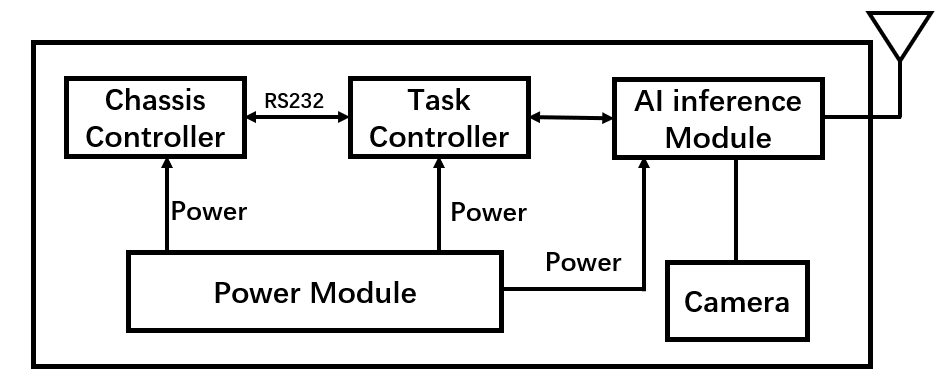}}
\centerline{(b) Unmanned vehicle structure}
\end{minipage}
\caption{The prototype hardware platform consists of edge servers and unmanned vehicles. The unmanned vehicle is equipped with a camera and an AI inference module.}
\label{Hardware_platform}
\end{figure}

The experimental hardware platform of the prototype is shown in Fig. \ref{Hardware_platform}. The edge server is an Intel NUC 11BTMi with a 3080 GPU. The structure of the unmanned vehicle is shown in Fig. \ref{Hardware_platform}(b), including a chassis controller, a task controller, an AI inference module, a power module, and a camera. NVIDIA Jetson AGX Orin is selected as the AI inference module, which has 40 TOPS AI computing power. 
The camera collects image data in real time, and the AI inference module uses the collected data as input data to complete the model inference. The task controller controls the vehicle speed according to the control commands from the AI inference module and sends the commands to the chassis controller through the RS232 serial port.

\subsection{Test Steps and Result Analysis}
The test process is described below,
\begin{itemize} 
\item Establish a basic network environment and set up wireless links(such as WiFi).
\item The edge server transmits the entire model for the new target to the AI inference module (device A) of unmanned vehicle 1, enabling device A to have the ability to detect new targets. At the same time, the edge server decouples the components of the new target detection model and transfers the model components to device B of the unmanned vehicle 2.
\item Capture data packets through the software Wireshark to test the amount of data received by device A and embedded device B.
\end{itemize}

\begin{figure}
\centering
\begin{minipage}{0.8\linewidth}
\centerline{\includegraphics[width=\textwidth]{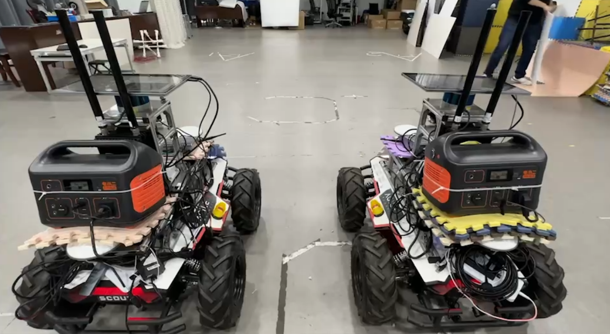}}
\centerline{(a) Two unmanned vehicles are at the same starting point.}
\end{minipage}
\begin{minipage}{0.8\linewidth}
\centerline{\includegraphics[width=\textwidth]{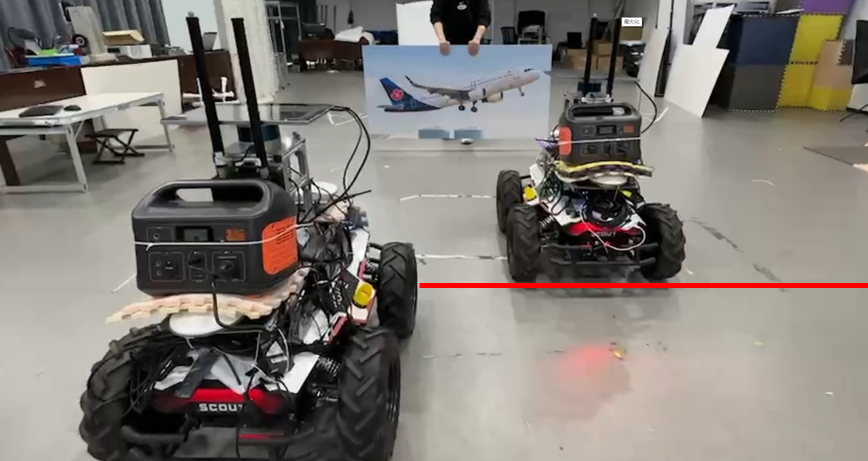}}
\centerline{(b) Scene where the first target appears.}
\end{minipage}
\begin{minipage}{0.8\linewidth}
\centerline{\includegraphics[width=\textwidth]{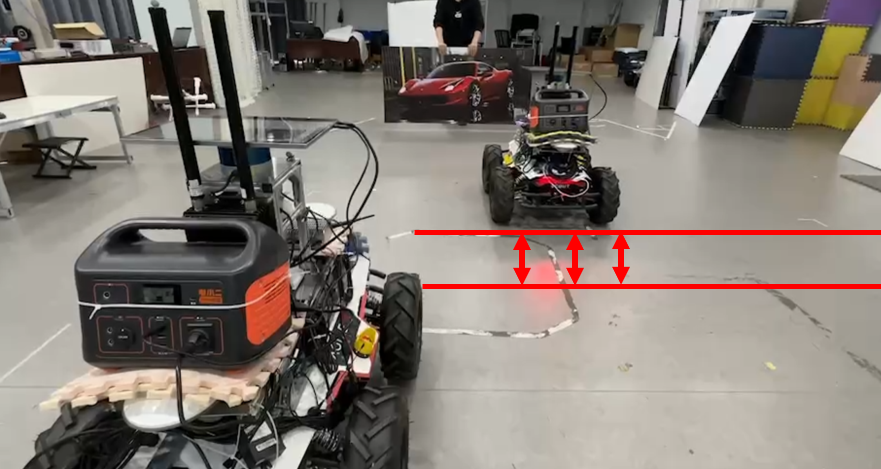}}
\centerline{(c) Scene where the second target appears.}
\end{minipage}
\begin{minipage}{0.8\linewidth}
\centerline{\includegraphics[width=\textwidth]{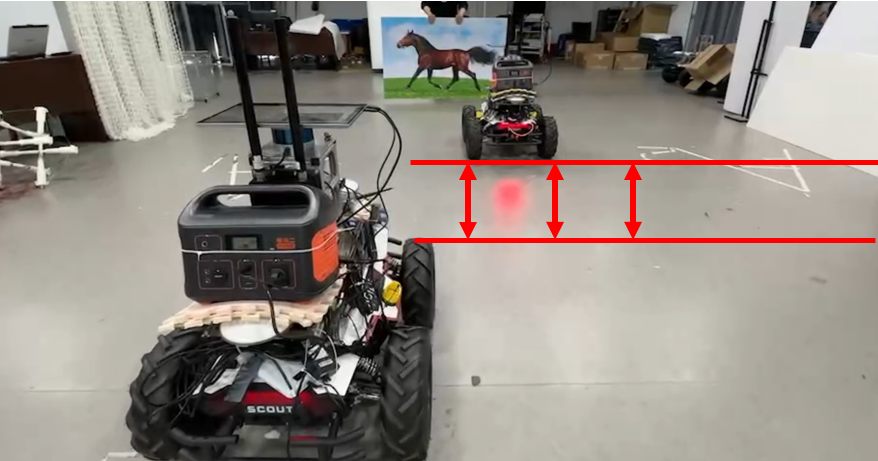}}
\centerline{(d) Scene where the third target appears.}
\end{minipage}
\caption{When the new unknown targets appear, unmanned vehicles using the component transmission scheme react faster and begin to track the target. As the number of unknown targets increases, the distance between the two vehicles gradually increases.}
\label{Scenario}
\end{figure}

In the unmanned vehicle identification new target experiment, the scene is shown in Fig. \ref{Scenario}. The three new targets are replaced by posters of aircraft, vehicles, and horses, with unmanned vehicle 1 on the left and unmanned vehicle 2 on the right. From Fig. \ref{Scenario}, it can be seen that each time a new target is changed, unmanned vehicle 2, using the SMC transmission scheme, will react more quickly and track the target.

Using transmission packet capturing software, we measured that the amount of data transferred by the edge server to the unmanned vehicle 1 and unmanned vehicle 2 was 129.8MB and 36.1MB, and the transmission time was 153s and 44s, respectively.

\section{Conclusion}
In semantic communication networks, edge nodes face multi-source and multi-task semantic extraction recovery services. Due to the limitation of storage capacity and computing power, edge nodes need the base station to send corresponding SMCs for parameter updates for different tasks and sources. Model distribution in the form of SMCs is performed to avoid performing all model parameter updates at the edge nodes when facing different scenarios. The base station weighs the model performance and transmission parameter size and then updates some parameters at the edge node through SMC form to save the model parameter transmission time, transmission bandwidth and model update time to some extent. In addition, we propose a method to improve the noise immunity of the model parameters during the propagation process. Finally, a component transfer based unmanned vehicle tracking prototype was implemented to verify the feasibility of model components in practical applications.

\section*{Acknowledgment}
This work is supported in part by the National Key R$\&$D Program of China under Grant 2022YFB2902102, in part by the National Natural Science Foundation of China under Grant 61871045 and Fundamental Research Funds for the Central Universities(Project Number: 2021RC01.)

\small
\bibliography{IEEEabrv, reference}

\end{document}